\pdfoutput=1

\documentclass[11pt]{article}

\usepackage[]{ACL2023}

\usepackage{times}
\usepackage{latexsym}

\usepackage[T1]{fontenc}

\usepackage[utf8]{inputenc}

\usepackage{microtype}

\usepackage{inconsolata}

\usepackage{xcolor}
\usepackage{lipsum}
\usepackage{booktabs}
\usepackage{multicol}
\usepackage{multirow}
\usepackage{graphicx}
\usepackage{tabularx}
\usepackage{float}
\usepackage{bm}
\usepackage{todonotes}
\usepackage{xargs}
\usepackage{xspace}
\usepackage{amsmath}
\usepackage{cleveref}
\usepackage{subcaption}
\usepackage{mathtools}
\usepackage{bigstrut}
\usepackage{tikz}
\usetikzlibrary{shapes}
\usepackage{rotating}
\usepackage{siunitx}
\usepackage{amssymb}
\usepackage{arydshln}
\usepackage{listings}
\usepackage[frozencache,cachedir=minted-cache,newfloat]{minted}
\usepackage{xfrac}

\usepackage{caption}

\usepackage{dcolumn}
\newcolumntype{L}{D{.}{.}{2,1}}

\newcommand*{\eg}{e.g.\@\xspace}
\newcommand*{\ie}{i.e.\@\xspace}

\makeatletter\newcommand*{\etc}{%
	\@ifnextchar{.}%
	{etc}%
	{etc.\@\xspace}%
}
\makeatother

\newcommand{\lde}{\textsc{De}\@\xspace}
\newcommand{\len}{\textsc{En}\@\xspace}
\newcommand{\les}{\textsc{Es}\@\xspace}
\newcommand{\lfr}{\textsc{Fr}\@\xspace}
\newcommand{\lit}{\textsc{It}\@\xspace}
\newcommand{\lja}{\textsc{Ja}\@\xspace}
\newcommand{\lko}{\textsc{Ko}\@\xspace}
\newcommand{\lpt}{\textsc{Pt}\@\xspace}
\newcommand{\lsw}{\textsc{Sw}\@\xspace}
\newcommand{\lzh}{\textsc{Zh}\@\xspace}

\newcommand{\lxen}{\textsc{X}$\rightarrow$\len}

\newcommand{\adam}{\textsc{Adam}\@\xspace}
\newcommand{\ours}{\textsc{LSL}\@\xspace}
\newcommand{\oursm}{{\ours}s\@\xspace}
\newcommand{\ourslong}{Language-Specific Transformer Layer\@\xspace}
\newcommand{\ourslongencoder}{Language-Specific Transformer Encoder Layer\@\xspace}
\newcommand{\sacrebleu}{\texttt{sacreBLEU}\xspace}
\newcommand{\chrf}{\textsc{chrF}\xspace}
\newcommand{\spBLEU}{\textsc{spBLEU}\@\xspace}
\newcommand{\comet}{\textsc{Comet}\@\xspace}
\newcommand{\improvement}{$1.5$\@\xspace}
\newcommand{\lsffn}{\textsc{LS-FFN}\@\xspace}
\newcommand{\lsattn}{\textsc{LS-Attention}\@\xspace}

\newcommand{\oursnas}{\ours-\textsc{Nas}\@\xspace}
\newcommand{\oursnassd}{\ours-\textsc{Nas-SD}\@\xspace}
\newcommand{\arch}[2]{\ours(\textsc{src}=$\{#1\}$ \& \textsc{tgt}=$\{#2\}$)}
\newcommand{\archemptysrc}[1]{\ours(\textsc{src}=$\emptyset$ \& \textsc{tgt}=$\{#1\}$)}
\newcommand{\archemptytgt}[1]{\ours(\textsc{src}=$\{#1\}$ \& \textsc{tgt}=$\emptyset$)}

\DeclareMathOperator*{\argmax}{arg\,max}

\DeclareMathOperator*{\softmax}{softmax}

\definecolor{red}{RGB}{141,45,57}
\definecolor{dark}{RGB}{55,65,74}
\definecolor{blue}{RGB}{0,105,170}
\definecolor{gold}{RGB}{174,159,109}
\definecolor{gray}{RGB}{175,179,183}
\definecolor{darkgreen}{RGB}{50,110,30}

\definecolor{ptgreen}{RGB}{213,232,212}

\newcommand{\coloredsquare}[1]{\raisebox{0.6pt}{\tikz{\node[draw,scale=0.4,regular polygon, regular polygon sides=4,fill=#1](){};}}}

\newcommand{\squarept}{\coloredsquare{ptgreen}}

\makeatletter
\newcommand*\standardbin{+}
\newcommand*\tabularbin[1]{%
  \mathbin{\mathpalette{\@tabularsym\standardbin}{#1}}%
}
\newcommand*\@tabularsym[3]{%
  \setbox\z@\hbox{$#2#1\m@th$}%
  \hbox to\wd\z@{\hss$#2#3\m@th$\hss}%
}
\makeatother

\newcolumntype{Y}{>{\centering\arraybackslash}X}

\newcommand{\chunb}[1]{$\hspace{-0.2cm}\textcolor{darkgreen}{\mathbf{\tabularbin+#1}}$} 

\newcommand{\insig}{\emph{non-significant}\xspace}

\newcommandx{\robin}[2][1=]{\vspace{0.2cm} \todo[linecolor=blue,backgroundcolor=blue!15,bordercolor=blue, #1]{\textbf{Robin:} #2}}
\newcommandx{\telmo}[2][1=]{\vspace{0.2cm} \todo[linecolor=gold,backgroundcolor=gold!25,bordercolor=gold, #1]{\textbf{Telmo:} #2}}
\newcommandx{\yihsiu}[2][1=]{\vspace{0.2cm} \todo[linecolor=darkgreen,backgroundcolor=darkgreen!25,bordercolor=darkgreen, #1]{\textbf{Yi-Hsiu:} #2}}

\newcommand{\todotemplate}{\textcolor{red}{TODO}\xspace}
\newcommandx{\trobin}[2][1=]{\vspace{0.2cm} \todo[linecolor=blue,backgroundcolor=blue!15,bordercolor=blue, #1]{\textbf{\todotemplate @Robin:} #2}}
\newcommandx{\ttelmo}[2][1=]{\vspace{0.2cm} \todo[linecolor=gold,backgroundcolor=gold!25,bordercolor=gold, #1]{\textbf{\todotemplate @Telmo:} #2}}
\newcommandx{\tstephan}[2][1=]{\vspace{0.2cm} \todo[linecolor=darkgreen,backgroundcolor=darkgreen!25,bordercolor=darkgreen, #1]{\textbf{\todotemplate @Stephan:} #2}}

\makeatletter
\def\adl@drawiv#1#2#3{%
        \hskip.5\tabcolsep
        \xleaders#3{#2.5\@tempdimb #1{1}#2.5\@tempdimb}%
                #2\z@ plus1fil minus1fil\relax
        \hskip.5\tabcolsep}
\newcommand{\cdashlinelr}[1]{%
  \noalign{\vskip\aboverulesep
           \global\let\@dashdrawstore\adl@draw
           \global\let\adl@draw\adl@drawiv}
  \cdashline{#1}
  \noalign{\global\let\adl@draw\@dashdrawstore
           \vskip\belowrulesep}}
\makeatother

\makeatletter
\newcommand{\ssymbol}[1]{^{\@fnsymbol{#1}}}
\newcommand{\nosig}{\ssymbol{2}\xspace}
\makeatother

\title{Learning Language-Specific Layers for Multilingual Machine Translation}

\author{Telmo Pessoa Pires \quad Robin M.~Schmidt \quad Yi-Hsiu Liao \quad Stephan Peitz \\
  Apple \\
  \texttt{\{telmo, robin\_schmidt, yihsiu\_liao, speitz\}@apple.com}}

\begin{document}
\maketitle
\begin{abstract}
Multilingual Machine Translation promises to improve translation quality between non-English languages. This is advantageous for several reasons, namely lower latency (no need to translate twice), and reduced error cascades (\eg, avoiding losing gender and formality information when translating through English).
On the downside, adding more languages reduces model capacity per language, which is usually countered by increasing the overall model size, making training harder and inference slower.
In this work, we introduce \ourslong{s} (\oursm), which allow us to increase model capacity, while keeping the amount of computation and the number of parameters used in the forward pass constant.
The key idea is to have some layers of the encoder be source or target language-specific, while keeping the remaining layers shared. We study the best way to place these layers using a neural architecture search inspired approach, and achieve an improvement of $1.3$ \chrf (\improvement \spBLEU{}) points over not using \oursm on a separate decoder architecture, and $1.9$ \chrf ($2.2$ \spBLEU{}) on a shared decoder one.\looseness=-1

\end{abstract}

\section{Introduction}
Multilingual Neural Machine Translation (MNMT) has received much attention from the Machine Translation community in recent years \cite{johnson-etal-2017-googles,aharoni-etal-2019-massively,freitag-firat-2020-complete,zhang-etal-2020-improving,JMLR:v22:20-1307,yang-etal-2021-multilingual-machine,tran-etal-2021-facebook}. This interest is based on the many advantages it provides:

\paragraph{Scalability}
Instead of having one model per language pair, a single model suffices, significantly reducing maintenance efforts as well as the combined model size across all languages.

\paragraph{Inference Speed and Less Error Cascading}
Due to the availability of data, most production systems are English-centric, meaning translation between two non-English languages na\"ively involves translating twice (\ie pivoting), once to English, and once from English. This approach increases latency and contributes to error cascades, since the translation to or from English usually implies information loss, \eg missing gender or formality distinctions that do not exist similarly in English.

\paragraph{Low-Resource Improvements}
Having a single model capable of handling multiple languages, means it can generalize across language boundaries and utilize characteristics of closely related translation directions to improve the translation quality for low-resource language-pairs (\ie knowledge transfer). Although achieving good zero-shot translation quality remains a challenging task, MNMT has been shown to help \citep{johnson-etal-2017-googles}.

\begin{figure*}[!ht]
    \centering
    \begin{subfigure}[t]{0.38\textwidth}
        \centering
        \includegraphics[width=0.45\textwidth]{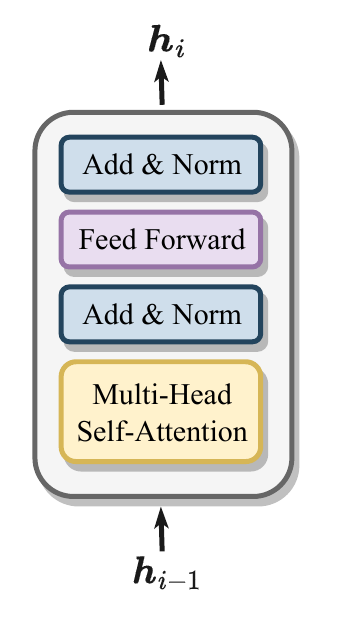}
        \caption{Regular Transformer Encoder Layer.}
        \label{fig:mtl:regular}
    \end{subfigure}
    \hfill
    \begin{subfigure}[t]{0.6\textwidth}
        \centering
        \includegraphics[width=0.8\textwidth]{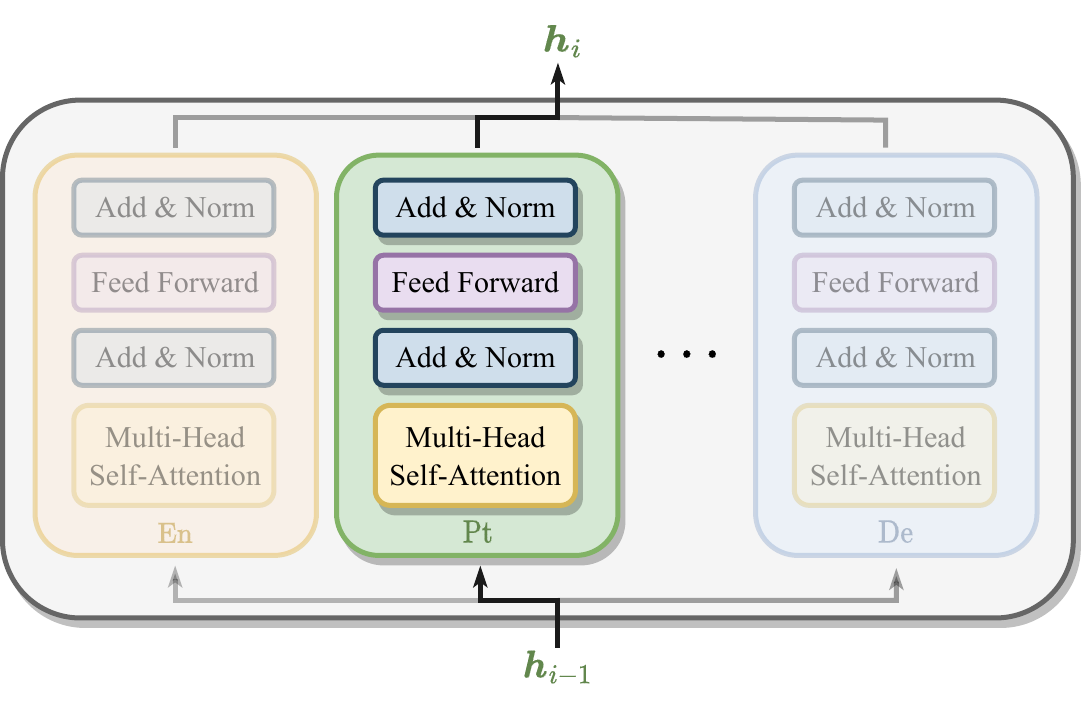}
        \caption{\ourslongencoder.}
        \label{fig:mtl:mtl}
    \end{subfigure}

    \caption{Side-by-side comparison of the regular Transformer Encoder Layer (a) and our \ourslongencoder (b). For the \ourslongencoder, an indexing language is provided which routes whole sentences to the appropriate weights (here \emph{Portuguese} \squarept).} 
    \label{fig:mtl}
\end{figure*}

\medskip\noindent Despite the above advantages, training high quality multilingual models is a challenging task: as more languages are added, the more they compete for the model's parameters \citep{sachan-neubig-2018-parameter}. A common solution is to increase the model size, but blindly doing so comes with its own troubles, as training becomes harder, inference slower, and the storage requirements increase, which makes them challenging to deploy to portable devices.

\begin{figure}[t]
    \centering
    \includegraphics[width=0.9\columnwidth]{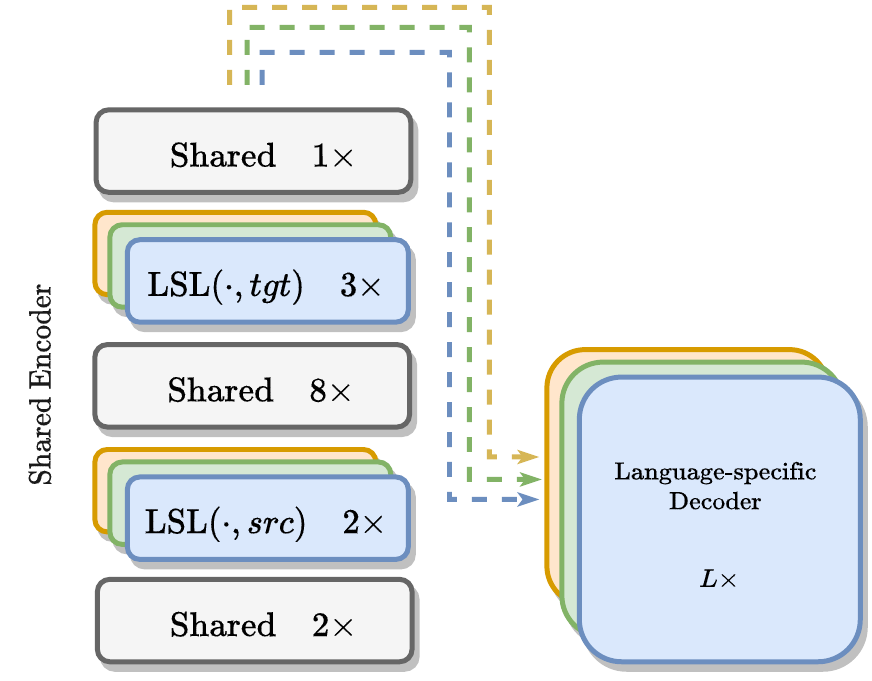}
    \caption{Best-performing separate decoder architecture using \oursm, found using our architecture search method. Layers $3$ and $4$ are indexed by the source language and layers $13$, $14$, and $15$ are indexed by the target language. The indices start at $1$ on the bottom of the encoder.}
    \label{fig:mtl-architecture}
\end{figure}

In this work, our goal is to increase the model capacity per language pair, while at the same time, letting the model share knowledge between languages, and without increasing the inference cost.
To this end, and combined with the observation from \citet{kudugunta-etal-2019-investigating} that the translation process in Transformer models starts in the top encoder layers, we propose an architecture with shared and language-specific weights. \Cref{fig:mtl-architecture} shows one such architecture, where layers $3$ and $4$\footnote{Throughout the paper, we use layer indices starting at $1$.} are source language-specific, layers $13$, $14$, and $15$ are target language-specific, the remaining layers are shared across all languages, and the decoder is also target language-specific. For the non-shared layers, we propose using \ourslong{s} (\oursm), illustrated in \Cref{fig:mtl:mtl}. Quite simply, \oursm are a combination (\ie, a dictionary) of regular Transformer layers (\Cref{fig:mtl:regular}), where the sub-layer used depends on the chosen language. We consider two cases: source-indexed \oursm, and target-indexed \oursm, distinguished by whether we use the source or the target language to select the appropriate sub-layer.

\medskip\noindent The main contributions of this work are:
\begin{enumerate}
    \item We propose a way to increase the model capacity per language, without changing the inference speed.
    \item We show that the model benefits from having both language-specific and shared components, as well as from having source and target language-specific components.
    \item We propose a technique to aid in learning the best architecture, rather than relying purely on manual trial-and-error.
\end{enumerate}

\section{Related Work}

There exists a vast literature investigating parameter sharing mechanisms for MNMT. Particularly relevant is the shared-encoder, separate-decoder architecture proposed by \citet{dong-etal-2015-multi} which we use as the base for some of our experiments.


Several works analyze which weights should be shared between languages \citep{sachan-neubig-2018-parameter, blackwood-etal-2018-multilingual, platanios-etal-2018-contextual, zhu-etal-2020-language, wang-etal-2019-compact, wang-etal-2018-three}. Regardless, most closely related to the presented work are the studies by \citet{ZhangBSF21} and \citet{purason-tattar-2022-multilingual}. \citet{ZhangBSF21} propose adding Conditional Language-Specific Routing (CLSR) layers inside the encoder and decoder Transformer layers. They learn to mix between language-specific and shared weights, and do this on a word by word basis. Our approach does not use learned routing but uses the same components for the whole sentence per language-pair, instead of computing a mixed representation. We also do not add extra parameters to the layer, meaning we have the same inference time complexity as regular Transformer layers. The approach in \citet{purason-tattar-2022-multilingual} is similar to ours in the sense that they use language-specific Transformer layers on the encoder side, and also look into sharing weights on a language-family basis. In contrast to our approach, they focus on source-indexed language-specific layers, while we investigate selecting the layers based on the source or the target language. Besides, we propose a systematic method for deciding which layers to share, and which to be language specific.

\paragraph{Connection to Adapter Layers}
\label{sec:adapters}
Adapter Layers \citep{HoulsbyGJMLGAG19,bapna-firat-2019-simple,he2022towards} are a lightweight technique to fine-tune a pre-trained encoder model by injecting task-specific sub-modules into the existing architecture. In contrast, \oursm are designed to be trained from scratch, and replace shared by language-specific components, rather than adding new ones, keeping the overall computational costs constant.

\paragraph{Connection to Mixture-of-Experts}
\label{sec:moe}
\oursm enable the introduction of source- and target-specific parameters in the encoder and increase the model capacity, while at the same time keeping the inference cost and effective parameter count for the forward-pass constant (see \Cref{fig:mtl}). As such, they are similar in nature to sparsely activated mixture-of-experts layers (MOEs, \citealp{ShazeerMMDLHD17,RollerSSW21,LepikhinLXCFHKS21}) but with the important differences that 1) there is no need for learning a balanced routing module;
2) sub-layer utilization is enforced by design, which tends to be a problem for MOE layers \citep{dua-etal-2022-tricks}; and 3) sentences are always routed to the same conditional compute based on the indexing-language, enabling smaller binaries for on-device downloading of model weights as well as consecutive downloads to extend the on-device capabilities to new languages. In fact, \citet{kudugunta-etal-2021-beyond-distillation} have shown that the final encoder MOE layers also learn target language-specific utilization where a subset of experts is used when translating \eg \lxen. However, since it is commonly \emph{not} strictly enforced, downloading \emph{all} experts is required, increasing the download size for end users.


\section{Methods}
In this section we describe our proposed \ourslong, as well as a way to select whether to use shared or language-specific weights for each layer.

\subsection{\ourslong}
\label{sec:lsl}
The idea of \oursm is simple: instead of sharing the same parameters across all languages, have the weights for the layer be language-specific as illustrated in \Cref{fig:mtl}. \oursm are composed of one ``regular'' Transformer encoder layer \emph{per language}. The input is routed to the appropriate sub-layer depending on the source or target language, and at any time only one of the sub-layers is used. Simply replacing all layers in the Transformer with \oursm would significantly increase the number of parameters, and reduce the sharing between languages. For example, if all \oursm are indexed by the source (or target) language it would be identical to a ``separate encoder separate decoder'' architecture. Instead, we propose a mix of \oursm and regular Transformer layers, which allows the model to learn language-specific and shared weights. See \Cref{fig:mtl-architecture} for one such architecture. A sample implementation for \textsc{fairseq} \citep{ott-etal-2019-fairseq} is given in \Cref{app:code_listing}.

\subsection{Learning the Architecture}
\label{sec:nas}
Intuitively, we expect the bottom layers of the encoder to require more source language knowledge, while the top ones should already capture target language information as found by \citet{kudugunta-etal-2019-investigating}. This observation motivates using source-indexed \oursm in the bottom encoder layers, target-indexed \oursm in the top ones, and keeping the remaining layers shared as illustrated in \Cref{fig:mtl-architecture}. This type of reasoning quickly gets out of hand, as the number of possible architectures is exponential in the numbers of layers. To avoid having to manually select which layers should be shared, and which should be source- or target-indexed \oursm, we propose a Neural Architecture Search \citep{JMLR:v20:18-598} inspired approach.

For each layer in the encoder, we learn a shared layer as well as one \ours, which can be source- and target-indexed, and $3$ scalar mixing weights:
\begin{align}
\label{eq:weighting}
    \boldsymbol{h}_{i} =\ &w_i^{shared} &\cdot \quad &\text{layer}_i^{shared}(\boldsymbol{h}_{i-1}) &+ \nonumber \\
    &w_i^{src} &\cdot \quad &\text{\ours}_i(\boldsymbol{h}_{i-1}, src) &+ \\
    &w_i^{tgt} &\cdot \quad &\text{\ours}_i(\boldsymbol{h}_{i-1}, tgt) & \nonumber,
\end{align}
where $\boldsymbol{h}_{i-1}$ and $\boldsymbol{h}_{i}$ are the outputs of layers $i-1$ and $i$, respectively, and $w_i^{shared} + w_i^{src} + w_i^{tgt} = 1$. $\text{\ours}_i(\boldsymbol{h}_{i-1}, src)$ means we select the \ours weights by the source language, while $\text{\ours}_i(\boldsymbol{h}_{i-1}, tgt)$ corresponds to using the target weights.

As there is no constraint on the mixing weights, other than that they sum to $1$ and are non-negative\footnote{We implement this constraint by applying the $\softmax$ function to the $3$ scalar parameters.}, the model is incentivized to use all the sub-layers, resulting in a huge increase in the number of parameters. If we have $L$ different languages, then each layer will have as many parameters as $L + 1$ ``regular'' Transformer layers.\footnote{Plus, of course, the mixing weights, but they amount to only $3$ extra parameters per layer.} The amount of computation increases by a factor of $3$, as we compute three intermediate representations: a shared one, one using the source language sub-layer, and another using the target language sub-layer, which we then mix according to \Cref{eq:weighting}.

To keep the inference time unaffected and the model size reasonable, only one of the components should be used, \ie, the mixing weights should be sparse. In this work, we propose a simple but effective approach: for each layer, we select the component with the largest converged weight. For example, if the largest weight for layer $i$ is $w_i^{tgt}$, then layer $i$ will be a target-indexed \ours. After selecting the architecture, we train it from scratch.

\subsection{Dense Pre-training}
\label{sec:dense}
Inspired by \citet{dua-etal-2022-tricks}, we found that initializing all encoder weights (both shared and \oursm) from a pre-trained architecture consisting only of ``regular'' Transformer layers helped achieve better performance.
In our experiments, we copy the pre-trained weights from the respective layers to the language-specific modules for initialization. The pre-trained weights come from our baseline architectures, shared-encoder models with only ``regular'' Transformer Layers. We use the separate decoder baseline's weights for the separate decoder models (\eg, \oursnas), and the shared decoder baseline's weights for the shared decoder models (\eg, \oursnassd).
This procedure has multiple advantages: 1) It maximizes cross-lingual transfer by training a general representation across languages first and minimizes language interference during fine-tuning; 2) It mitigates under-trained language-specific components for low-resource languages as they usually see significantly less data and the na\"ive approach of training with higher sampling temperatures typically degrades performance on high resource languages \citep{DBLP:journals/corr/abs-1907-05019,wang-etal-2020-balancing}; and 3) it improves convergence speed for architectures with \oursm.

\section{Results}
In the following, we will describe our experiments and discussion regarding the effectiveness of \oursm.

\subsection{Experimental Setup}

\paragraph{Data}
In our experiments, we focus on the following $10$ languages: German (\lde), English (\len), Spanish (\les), French (\lfr), Italian (\lit), Japanese (\lja), Korean (\lko), Portuguese (\lpt), Swahili (\lsw), and Chinese (\lzh). We collect data for these languages from the WMT21 news translation task sources (composed of Europarl v10, ParaCrawl v7.1, ParaCrawl v8, Common Crawl, News Commentary v16, Wiki Titles v3, UN Parallel Corpus V1.0, Tilde Rapid, WikiMatrix, Back-translated news, Japanese-English Subtitle Corpus, The Kyoto Free Translation Task Corpus, and TED Talks) as well as Opus-100 \citep{zhang-etal-2020-improving}, Tatoeba \citep{TIEDEMANN12.463}, and CCMatrix \citep{schwenk-etal-2021-ccmatrix}. We deduplicate the data, and preprocess it using the M2M-100 \citep{JMLR:v22:20-1307} scripts.\footnote{\url{https://github.com/facebookresearch/fairseq/tree/main/examples/m2m_100}} The final dataset sizes can be seen in \Cref{app:dataset_sizes}.

Since CCMatrix is a large yet low quality data source, we found it helpful to downsample it relative to the other sources using temperature sampling. For more details, see \Cref{app:dataset_sizes}.

\paragraph{Evaluation}
For evaluation, we use the dev and devtest splits of the Flores-101 dataset \citep{goyal-etal-2022-flores} as our validation and test sets, respectively. Except when stated otherwise, the reported numbers are on the test set. We report both \chrf \citep{popovic-2015-chrf} and \spBLEU \citep{goyal-etal-2022-flores},
a \textsc{SentencePiece}-based BLEU computed using the Flores-101 tokenizer, with \sacrebleu\footnote{\url{https://github.com/mjpost/sacrebleu}} version $2.3.1$. The evaluation signatures are \texttt{nrefs:1} \texttt{|} \texttt{case:mixed} \texttt{|} \texttt{eff:no} \texttt{|} \texttt{tok:flores101} \texttt{|} \texttt{smooth:exp} for \spBLEU, and \texttt{nrefs:1} \texttt{|} \texttt{case:mixed} \texttt{|} \texttt{eff:yes} \texttt{|} \texttt{nc:6} \texttt{|} \texttt{nw:0} \texttt{|} \texttt{space:no} for \chrf. All our results are from a single training run of each architecture, and we perform statistical significance tests using paired bootstrap resampling \citep{koehn-2004-statistical}. We run the significance tests for \chrf for all language directions, using a significance level of $5\%$. We also provide \comet scores \citep{rei-etal-2020-comet}\footnote{Obtained with \texttt{wmt20-comet-da} from version \texttt{1.1.2}.} for selected models in \Cref{app:comet_results}.

\paragraph{Tokenization}
We use \textsc{SentencePiece} \citep{kudo-richardson-2018-sentencepiece}, with a vocabulary size of $250$k, and a character coverage of $0.9995$. We balance the data for \textsc{SentencePiece} training by randomly sampling $1.5$M sentences per language.

\paragraph{Tagging}
We found it helpful to make the model aware of the corpus by training with corpus labels. Similarly to \citet{DBLP:journals/corr/abs-2207-04672}, we add a tag (\eg \texttt{<HQ>} or \texttt{<LQ>}) to the beginning of the source sentence, so that the model can learn to distinguish between higher quality (WMT21, Opus-100, and Tatoeba) and lower quality examples (CCMatrix).
During inference, we always use the high quality (\texttt{<HQ>}) tag. Additionally, we append source and target language tags to the end of the sentence.

\paragraph{Architecture}
In our experiments, we use a deep encoder, shallow decoder architecture \citep{Kasai0PCS21} with $16$ encoder layers and $3$ decoder layers. We share token embeddings between the encoder, decoder, and output layer \cite{press-wolf-2017-using}. In our experiments we consider two kinds of models: those with target language-specific decoders, following \citet{dong-etal-2015-multi}, on which we conduct most of our experiments, and those with a shared decoder. The encoder is always shared, with the exception of the \oursm. In the baseline models, the encoder consists only of ``regular'' Transformer Layers, and so it is fully shared.
In this work, we only consider adding \oursm to the encoder. In preliminary experiments with \oursm in the decoder, our selection criteria picked target-specific \oursm for all decoder layers, effectively choosing a separate decoder architecture. We tried different placements of the layers in the decoder, but did not achieve any improvements. We leave a deeper analysis to future work.


\paragraph{Hyperparameters}
All experiments are implemented using \textsc{fairseq} \citep{ott-etal-2019-fairseq}. We use \adam \citep{KingmaB14} for optimization, due to its robustness \citep{SchmidtSH21} and popularity, with a learning rate of $0.0004$. We train for $150$k steps, by which point our models had converged, with $4000$ warm-up steps, and an inverse square root learning rate scheduler \citep{NIPS2017_attention}. Due to the abundance of data, adding regularization in the form of dropout or weight decay did not help in our initial experiments, so we do not use any regularization in the remaining experiments. The layer and embedding sizes are $512$, the hidden size of the feed-forward layers is $2048$, and we use $8$ attention heads. All models are trained using \texttt{fp16} \citep{ott-etal-2018-scaling}.

\subsection{Architecture Search}
\label{sec:nas_results}

\begin{figure}[t]
    \centering
    \includegraphics[width=\columnwidth]{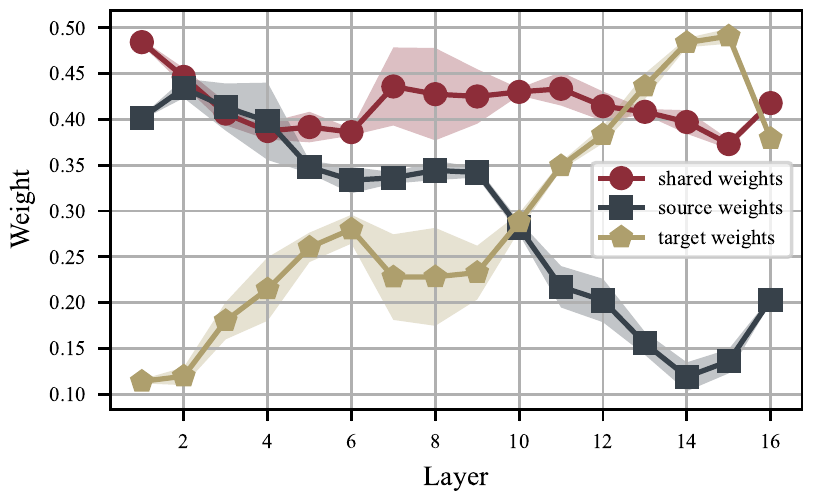}
    \caption{Converged mixing weights across layers, averaged over $3$ runs. The model shows a preference for source \oursm near the bottom of the encoder, target \oursm near the top, and shared layers in between. Shaded regions show the uncertainty.}
    \label{fig:converged-mixing-weights}
\end{figure}

As described in \Cref{sec:nas}, we train a separate-decoder model where all encoder layers are a mix of shared, source, and target weights. This architecture used a total of $804$ million (M) parameters, and achieved a score of $46.6$ \chrf points ($27.4$ \spBLEU), averaged over all language pairs. We plot the mixing coefficients of the model in \Cref{fig:converged-mixing-weights}, averaged over $3$ runs.

We can see clear trends here: the model gives a higher weight to the source-specific sub-layers near the bottom of the encoder, while the target-specific sub-layers get a higher weight near the top. This is in line with previous studies as lower encoder layers
usually capture low-level information about the source \citep{tenney-etal-2019-bert}, while the top encoder layers are known to already capture target language information \citep{kudugunta-etal-2019-investigating}. Interestingly, the mixing coefficients for the shared weights are relatively stable across layers, making them dominant for the middle layers of the model.



Taking the $\argmax$ of the mixing coefficients, we select the architecture in \Cref{fig:mtl-architecture}, where layers $3$ and $4$ are source-indexed \oursm\footnote{For these layers there is some uncertainty in the source weights, but they are the largest weights by a small margin. Performance is improved by selecting source layers, as can be attested by comparing to \archemptysrc{13,14,15}.}, layers $13$, $14$, $15$ are target-indexed \oursm, and the remaining layers are ``regular'' Transformer encoder layers (\Cref{fig:mtl:regular}). From here onward, we will refer to this architecture as \oursnas{}. We use the architecture selection method only to select the architecture, and the selected architecture is trained from scratch (\textbf{not pruned}) in the upcoming experiments. To simplify the text, we will also use the notation \arch{1,2}{15,16} to refer to an architecture with source-indexed \oursm in layers $1$ and $2$, and target-indexed \oursm in layers $15$ and $16$.

\subsection{Learned Architecture Comparison}
\label{sec:arch_comparison}
In \Cref{tab:sep_dec_results}, we compare our baseline separate decoder architecture (with a fully shared $16$ layer encoder) with the learned architecture from the architecture search (\oursnas), and additional variants. We share \chrf and \spBLEU{} scores averaged over all language pairs, as well as the number of total ($|\boldsymbol{\theta}|$) and effective ($|\boldsymbol{\theta}_{\text{eff}}|$) parameters used during inference for each architecture. For the baseline models, $|\boldsymbol{\theta}|$ and $|\boldsymbol{\theta}_{\text{eff}}|$ differ due to the separate decoders. For an accurate comparison of CPU and GPU speed, see \Cref{app:inference_speed}.

\begin{table}[t]
    \centering
    \scriptsize
    {\setlength{\tabcolsep}{1.2pt}
    \begin{tabular}{lcccc}
    \toprule
    \multicolumn{1}{l}{Model} & \chrf & \spBLEU & $|\boldsymbol{\theta}|$ & $|\boldsymbol{\theta}_{\text{eff}}|$ \\
    \midrule
    \multicolumn{1}{l}{Separate Decoder Baseline} & $45.5$ & $26.0$ & $299$M & $186$M \\
    \multicolumn{1}{l}{\quad + hidden dim $640$} & $45.9$ & $26.6$ & $399$M & $240$M \\
    \multicolumn{1}{l}{\quad + hidden dim $704$} & $46.2$ & $26.9$ & $453$M & $268$M \\
    \multicolumn{1}{l}{\quad + hidden dim $768$} & $46.5$ & $27.3$ & $509$M & $296$M \\
    \midrule
    Language Adapters \textsc{Enc} $128$ & $45.8$ & $26.4$ & $321$M & $207$M \\
    Language Adapters \textsc{Enc} $256$ & $45.7$ & $26.3$ & $342$M & $228$M \\
    Language Adapters \textsc{Enc} $512$ & $45.6$ & $26.3$ & $384$M & $270$M \\
    \cdashlinelr{1-5}
    Language Adapters \textsc{Enc (Src+Tgt)} $128$ & $45.7$ & $26.3$ & $321$M & $207$M \\
    Language Adapters \textsc{Enc (Src+Tgt)} $256$ & $46.1$ & $26.7$ & $342$M & $228$M \\
    Language Adapters \textsc{Enc (Src+Tgt)} $512$ & $46.0$ & $26.7$ & $384$M & $270$M \\
    \midrule
    \multicolumn{1}{l}{\oursnas} & $46.4$ & $27.2$ & $441$M & $186$M \\
    \multicolumn{1}{l}{\quad \emph{+ Dense Pre-training}} & $\mathbf{46.8}$ & $\mathbf{27.5}$ & $441$M & $186$M \\
    \arch{1,2}{15,16} & $46.3$ & $27.0$ & $413$M & $186$M \\
    \arch{1,2,3}{14,15,16} & $46.2$ & $27.0$ & $470$M & $186$M \\
    \cdashlinelr{1-5}
    \archemptysrc{13,14,15} & $45.8$ & $26.5$ & $385$M & $186$M \\
    \archemptytgt{3,4} & $46.1$ & $26.7$ & $356$M & $186$M \\
    \cdashlinelr{1-5}
    \arch{13,14,15}{3,4} & $45.2$ & $25.7$ & $441$M & $186$M \\
    \bottomrule
    \end{tabular}}
    \caption{Comparison of different separate decoder models. Although the total number of parameters in the model $|\boldsymbol{\theta}|$ changes by adding more \oursm, the effective parameter count $|\boldsymbol{\theta}_{\text{eff}}|$ stays consistent for all translation directions due to sparse language-dependent activations.}
    \label{tab:sep_dec_results}
\end{table}
\begin{table*}[t]
\centering
\scriptsize
\begin{tabularx}{\textwidth}{lYYYYYYYYYY}
\toprule
Model & \lde & \len & \les & \lfr & \lit & \lja & \lko & \lpt & \lsw & \lzh \\
\midrule
\multicolumn{11}{c}{Translating \emph{into} the language ($\mathrm{X}$ $\rightarrow$ \underline{\hspace{0.3cm}} )} \\
\midrule
Separate Decoder Baseline & $52.7$ & $60.4$ & $49.1$ & $57.0$ & $50.6$ & $29.0$ & $25.2$ & $54.9$ & $47.9$ & $28.7$ \\
Language Adapters \textsc{Enc (Src+Tgt)} $256$ & $53.2$ & $60.9$ & $49.5$ & $57.7$ & $50.9$ & $29.9$ & $25.0$ & $55.4$ & $49.1$ & $29.1$ \\
\oursnas & $\mathbf{53.9}$ & $\mathbf{61.5}$ & $\mathbf{49.9}$ & $\mathbf{58.1}$ & $\mathbf{51.4}$ & $\mathbf{31.0}$ & $\mathbf{27.0}$ & $\mathbf{55.6}$ & $\mathbf{49.4}$ & $\mathbf{29.9}$ \\

\midrule
\multicolumn{11}{c}{Translating \emph{from} the language ( \underline{\hspace{0.3cm}} $\rightarrow$ $\mathrm{X}$)} \\
\midrule
Separate Decoder Baseline & $47.7$ & $52.7$ & $44.9$ & $48.1$ & $46.4$ & $42.1$ & $40.0$ & $49.4$ & $40.1$ & $44.0$ \\
Language Adapters \textsc{Enc (Src+Tgt)} $256$ & $48.1$ & $53.0$ & $45.4$ & $48.5$ & $46.7$ & $42.8$ & $41.2$ & $49.9$ & $40.8$ & $44.5$ \\
\oursnas & $\mathbf{48.7}$ & $\mathbf{53.6}$ & $\mathbf{45.8}$ & $\mathbf{48.9}$ & $\mathbf{47.3}$ & $\mathbf{43.5}$ & $\mathbf{42.1}$ & $\mathbf{50.4}$ & $\mathbf{42.1}$ & $\mathbf{45.3}$ \\

\bottomrule
\end{tabularx}
\caption{Comparison of \oursnas with pre-training, the separate decoder baseline model and the best separate decoder adapter model, per source and target language. Our approach gives substantial average \chrf improvements over the baseline (adapter) model, which are statistically significant for $84$ ($62$) of the $90$ translation directions.}
\label{tab:fine-grained-results}
\end{table*}


Our learned architecture (\oursnas{} in \Cref{tab:sep_dec_results}) achieves a $0.9$ \chrf ($1.2$ \spBLEU) improvement over the baseline, which we can be further increased to $1.3$ with dense pre-training, reaching a total of $46.8$ \chrf ($27.5$ \spBLEU). These improvements are statistically significant ($p < 0.05$) for all but $6$ of the $90$ translation directions. In \Cref{tab:fine-grained-results}, we summarize the averaged results for translating to and from each language, \ie X$\rightarrow$ \lde is the average \chrf score for translating into German from all other languages. For the full results (per language pair) on the validation and test sets, see \Cref{app:full_results}.
Our approach gives substantial gains for both high resource languages, such as English and German, which improve by more than $1$ \chrf point, as well as lower resource, such as Korean, with close to $2$ \chrf points improvement for both directions, or Swahili, which improves by over $1.5$ \chrf points in both directions.
Although the effective number of parameters is the same for this architecture and our baseline ($186$M), it can be argued that this comparison is unfair, since our model is bigger. To alleviate this concern, and to show that the gains we achieve are not just due to the higher parameter count, but rather, the better way we allocate the extra parameters, we trained three bigger baselines: with hidden sizes of $640$, $704$, and $768$. As expected, these models also show an improvement over the original baseline, but even the biggest model, with a total of $509$M parameters ($15\%$ more than ours) and a higher inference cost than our method, is not able to match our performance (only $46.5$ \chrf and $27.3$ \spBLEU).

\paragraph{Adapters} Following \citet{philip-etal-2020-monolingual}, we insert one adapter block after each Transformer layer. In our experiments, inserting adapters into a pre-trained model either provided no improvement over training from scratch or suffered from numerical instabilities, even after tuning the initialization gain \citep{HoulsbyGJMLGAG19}.  For this reason we report numbers for models trained from scratch, similar to \citet{Baziotis2022MultilingualMT}. Since our models have separate decoders, we inserted adapters only on the encoder. For completeness, results using adapters on the decoder are reported in \Cref{app:different_configs}.

We consider two kinds of adapters: source language adapters (Language Adapters \textsc{Enc}), following \citet{philip-etal-2020-monolingual}, or source language adapters in the bottom half of the encoder and target language language adapters in the top half (Language Adapters \textsc{Enc (Src+Tgt)}). We show the results for different bottleneck dimensions ($128$, $256$, and $512$) in \Cref{tab:sep_dec_results}. Our proposal of using source and target adapters on the encoder outperforms using only source adapters (for the same model size). The best performing model, Language Adapters \textsc{Enc (Src+Tgt)}, achieves a score of $46.1$ \chrf points, $0.3$ ($0.7$) points lower than our model without (with) dense pre-training.
These improvements are statistically significant ($p < 0.05$) for $38$ ($62$) of the $90$ translation directions. Results for language-pair adapters \citep{bapna-firat-2019-simple} are shown in \Cref{app:different_configs}, but they lag behind language adapters.


\paragraph{Importance of Bottom and Top Shared Layers}
\oursnas{} uses two shared layers on the bottom and one shared layer on the top of the encoder. In \Cref{tab:influence_of_top_bottom_shared}, we analyze the effect of removing these layers, \ie, moving the \oursm up or down.
When comparing \spBLEU{} there is a small drop when removing either the top shared layer (row ``\arch{3,4}{14,15,16}'') or the bottom-most shared layer (row ``\arch{2,3}{13,14,15}''), but the difference is negligible when comparing \chrf. In fact the difference is only statistically significant for $15$ of the $90$ translation directions.
When removing the bottom shared layers (row ``\arch{1,2}{13,14,15}'') or all the shared layers (row ``\arch{1,2}{14,15,16}''), there is a bigger difference, but it is only statistically significant for less than $\sfrac{1}{3}$ of the translation directions, mostly low resource pairs including either Swahili or Korean. For an analysis regarding the number of \oursm, please refer to \Cref{app:number_of_lsls}.


\begin{table}[t]
    \centering
    \scriptsize
    {\setlength{\tabcolsep}{3pt}
    \begin{tabular}{lcc}
    \toprule
    Model & \chrf & \spBLEU \\
    \midrule
    \oursnas & $\mathbf{46.4}$ & $\mathbf{27.2}$ \\
    \arch{3,4}{14,15,16} & $46.4$ & $27.0$ \\
    \arch{2,3}{13,14,15} & $46.4$ & $27.1$ \\
    \arch{1,2}{13,14,15} & $46.2$ & $26.9$ \\
    \arch{1,2}{14,15,16} & $46.2$ & $26.9$ \\
    \bottomrule
    \end{tabular}}
    \caption{Influence of the shared layers on the bottom and the top of the encoder. Our learned architecture, \oursnas, is \arch{3,4}{13,14,15}.
    }
    \label{tab:influence_of_top_bottom_shared}
\end{table}

\paragraph{Alternative Configurations}
Additionally, we look at different configurations of \oursm. In particular, we compare using only source-specific layers \archemptytgt{3,4} or target-specific layers \archemptysrc{13,14,15} in \Cref{tab:sep_dec_results}. In both cases, the configuration is worse than \oursnas, thus showing the importance of having both source and target-specific layers. For completeness, row \arch{13,14,15}{3,4} shows the opposite of our configuration (\ie, swapping the source and target layers), with considerably degraded performance, showing that the position of source-specific and target-specific languages is very important. In particular, it shows that forcing the model to learn source-specific representations at higher encoder layers and target language representations on the lower layers hinders learning.

\begin{table}[t]
    \centering
    \scriptsize
    {\setlength{\tabcolsep}{3pt}
    \begin{tabular}{lcccc}
    \toprule
    Model & \chrf & \spBLEU & $|\boldsymbol{\theta}|$ & $|\boldsymbol{\theta}_{\text{eff}}|$ \\
    \midrule
    \oursnas & $\mathbf{46.4}$ & $\mathbf{27.2}$ & $441$M & $186$M \\
    \lsffn & $46.3$ & $26.8$ & $394$M & $186$M \\
    \lsattn & $45.9$ & $26.5$ & $347$M & $186$M \\
    \bottomrule
    \end{tabular}}
    \caption{Effect of each Transformer layer component. \ours uses full language-specific layers. \lsffn shares the attention, but keeps the feed-forwards language specific, while \lsattn does the opposite. All experiments are based on the \oursnas{} architecture and differ only in their language-specific components.}
    \label{tab:component_ablation}
\end{table}

\paragraph{Layer Component Ablation}
We analyze the effect of using full language-specific layers (\ours), with having only language-specific feed-forward (\lsffn), or only language-specific attention (\lsattn) on the \oursnas{} architecture in \Cref{tab:component_ablation}. We observe a small degradation of $0.1$ \chrf ($0.4$ \spBLEU) when switching from \ours to \lsffn, which is statistically significant for $35/90$ translation directions, and a degradation of $0.5$ \chrf ($0.7$ \spBLEU) when switching to \lsattn, which is significant for $49$ directions.
These results imply that both the language-specific feed-forward and attention are important, with the biggest contribution coming from the feed-forward part, where most of the parameters are located.

\subsection{Shared Decoder}
So far we have focused on the separate-decoder architecture. In this section, we turn to a shared-decoder setup (see \Cref{tab:results_shareddec}). As in \Cref{sec:nas_results}, we ran an architecture search experiment and selected the following architecture: \arch{4}{12,13,14,15,16}, or \oursnassd for short. The mixing weights follow a trend similar to \Cref{fig:converged-mixing-weights}.
With the shared decoder, we benefit from placing more target-specific layers at the top of the encoder. Our intuition is that these layers compensate the lack of a separate decoder.

As in \Cref{sec:arch_comparison}, we compare against shared decoder baseline models (\ie, without \oursm) of increasing sizes, as well as models with Adapter Blocks. For the latter, we insert one block after each Transformer layer, both on the encoder and the decoder. Following \citet{philip-etal-2020-monolingual}, we insert source adapters on the encoder, and target adapters on the decoder. As expected, shared-decoder models perform worse than their separate-decoder models, which have a higher parameter count. Despite this, our proposed architecture, \oursnassd, outperforms the remaining models by a wide margin, and is even better than the separate-decoder baseline (26.0 \spBLEU{}). The improvements of our \oursnassd{} model with pre-training over the shared decoder baseline are statistically significant for $86/90$ translation directions. The improvements over the best adapter model (bottleneck size $512$) are significant for $76/90$ directions.

We also show the performance for \arch{4}{13-16}, an architecture similar to \oursnassd, but with one less target-specific \ours. This architecture performs worse than our selection, but has fewer parameters, which might make it a preferable candidate for deployment. This highlights a limitation of our selection approach: it does not take model complexity (\ie, model size) into account. We tried adding a prior on the mixing weights to make \oursm more costly than shared layers, but obtained mixed results, and we leave further investigation to future work.

\begin{table}[t]
    \centering
    \scriptsize
    {\setlength{\tabcolsep}{1.2pt}
    \begin{tabular}{lcccc}
    \toprule
    \multicolumn{1}{l}{Model} & \chrf & \spBLEU & $|\boldsymbol{\theta}|$ & $|\boldsymbol{\theta}_{\text{eff}}|$ \\
    \midrule
    \multicolumn{1}{l}{Separate Decoder Baseline} & $45.5$ & $26.0$ & $299$M & $186$M \\
    \multicolumn{1}{l}{\oursnas (separate decoder)} & $46.4$ & $27.2$ & $441$M & $186$M \\
    \cdashlinelr{1-5}
    \multicolumn{1}{l}{Shared Decoder Baseline} & $44.7$ & $24.9$ & $186$M & $186$M \\
    \multicolumn{1}{l}{\quad + hidden dim $640$} & $45.1$ & $25.5$ & $240$M & $240$M \\
    \multicolumn{1}{l}{\quad + hidden dim $704$} & $45.8$ & $26.2$ & $268$M & $268$M \\
    \multicolumn{1}{l}{\quad + hidden dim $768$} & $45.8$ & $26.3$ & $296$M & $296$M \\
    \midrule
    Shared Decoder Adapters $128$ & $44.6$ & $24.8$ & $211$M & $189$M \\
    Shared Decoder Adapters $256$ & $44.9$ & $25.0$ & $236$M & $191$M \\
    Shared Decoder Adapters $512$ & $45.3$ & $25.6$ & $286$M & $196$M \\
    Shared Decoder Adapters $640$ & $45.3$ & $25.5$ & $311$M & $199$M \\
    \midrule
    \multicolumn{1}{l}{\oursnassd} & $46.3$ & $26.7$ & $356$M & $186$M \\
    \multicolumn{1}{l}{\quad \emph{+ Dense Pre-training}} & $\mathbf{46.6}$ & $\mathbf{27.1}$ & $356$M & $186$M  \\
    \multicolumn{1}{l}{\arch{4}{13-16}} & $46.1$ & $26.5$ & $328$M & $186$M \\
    \bottomrule
    \end{tabular}}
    \caption{Results on the shared-decoder architecture.}
    \label{tab:results_shareddec}
\end{table}

\subsection{Zero-shot Translation}
In the previous experiments, we used training data for all language directions. We now consider a different scenario: we limit our training data to English directions (\ie, X-\len and \len-X) and languages in the same language group\footnote{We consider $3$ groups: European, CJK, and Swahili. We use data where both the source and target languages are in the same group.}. We then evaluate our models on zero shot performance for the directions between groups.

In our initial experiments, separate decoder models performed poorly on zero-shot directions, so we focused our evaluation on shared decoder models. \Cref{tab:results_zeroshot} shows the zero-shot results for $3$ architectures: the shared decoder baseline, the best performing (shared decoder) adapter model (Shared Decoder Adapters $512$), and \oursnassd. Our approach gives improvements for most zero-shot directions, except when translating \emph{into} \lsw. Translating \emph{from} \lsw works well, though. Our intuition is that this degradation is caused by the \lsw target-specific \oursm being overfitted to \len, and thus failing to transfer to other languages. In \oursnassd{}, the top $5$ encoder layers are target \oursm, and in the zero-shot scenario, the \lsw layers are only trained for \len-\lsw, which is relatively small. Indeed, if we exclude the $\textsc{*}\rightarrow$\lsw pairs, both the overall and the zero-shot average scores increase.

\begin{table}[t]
    \centering
		\scriptsize
        {\setlength{\tabcolsep}{2pt}
		\begin{tabular*}{0.9\columnwidth}{lllccc}
    		\toprule
    		\multicolumn{3}{c}{Direction} & Baseline & Adapters & \oursnassd \\
    		\midrule
            \multicolumn{3}{l}{Overall Average} & $39.9$ & $\mathbf{41.8}$ & $41.4$ \\
            \multicolumn{3}{l}{Overall Average (w/o \textsc{*}$\rightarrow$\lsw)} & $40.8$ & $42.5$ & $\mathbf{44.7}$ \\
            \cdashlinelr{1-6}
            \multicolumn{3}{l}{Zero-shot Average} & $29.6$ & $\mathbf{32.4}$ & $31.9$ \\
            \multicolumn{3}{l}{Zero-shot Average (w/o \textsc{*}$\rightarrow$\lsw)} & $29.3$ & $32.0$ & $\mathbf{36.2}$ \\
            \midrule
            \textsc{Eur} & $\rightarrow$ & \textsc{CJK} & $23.8$ & $18.5$ & $\mathbf{27.6}$ \\
            \textsc{Eur} & $\rightarrow$ & \lsw & $34.2$ & $\mathbf{37.3}$ & $11.7$ \\
            \midrule
            \textsc{CJK} & $\rightarrow$ & \textsc{Eur} & $41.4$ & $45.1$ & $\mathbf{45.5}$ \\
            \textsc{CJK} & $\rightarrow$ & \lsw & $24.8$ & $\mathbf{28.9}$ & $11.7$ \\
            \midrule
            \lsw & $\rightarrow$ & \textsc{Eur} & $23.5$ & $44.9$ & $\mathbf{46.4}$ \\
            \lsw & $\rightarrow$ & \textsc{CJK} & $6.70$ & $12.3$ & $\mathbf{18.7}$ \\

    		\bottomrule
		\end{tabular*}}
		\caption{Zero-Shot comparison of the shared decoder models. The global average includes non zero-shot directions. The remaining scores are all zero-shot.}
		\label{tab:results_zeroshot}
\end{table}

\section{Conclusion}
In this work, we studied how to increase the capacity of MNMT models using \oursm. We showed that \oursm are effective at increasing the model capacity per language, while keeping the computation requirements constant.
We proposed a method for selecting the placement of \oursm, and showed the importance of having shared as well as source and target language-specific parameters on the encoder.


\section*{Limitations}
In this work, we focused our exploration of \oursm on the encoder. Although we ran some initial explorations on the decoder side, further investigation is needed.
Another venue for research is how \oursm affect language expansion. Since our approach tries to limit the language-specific weights to just a few layers, \emph{in theory}, it should be possible to add new languages by only expanding and training the \oursm. However, blindly doing so might not work well and the interactions between languages from different families needs further studying. Lastly, it is unclear whether our $\argmax$ approach to selecting where to place \oursm is optimal, how dataset dependent it is, and if there exist alternative approaches that can lead to better results. The fact that it does not take model complexity (\ie, model size) into account can be a disadvantage in practice.

\section*{Ethics Statement}
Our work uses existing datasets, so it inherits some of the risks associated with them,
namely gender bias \citep{cho-etal-2019-measuring}, or privacy leakage \citep{privacy-leakage-lm}, and mitigation strategies such as \citet{vanmassenhove-etal-2018-getting} may be necessary. However, replacing bilingual translation systems with multilingual systems should help reduce gender bias caused by pivoting through English.
Another consideration is the energy consumption for model training, which results in green-house emissions \citep{strubell-etal-2019-energy}. Our proposed architectures result in smaller (and faster to train) models, than similarly-performing baselines, increasing the efficiency of translation systems.

\section*{Acknowledgements}
We would like to thank Sarthak Garg, Luke Carlson, António V. Lopes, and Matthias Sperber for their comments and suggestions, which significantly improved the final work.

\newpage
\clearpage
\appendix

\section{\oursm in \textsc{fairseq}}
\label{app:code_listing}
\Cref{code:mtl} shows our implementation of \ours in \textsc{fairseq}. The implementation is straightforward, and consists of a dictionary that selects the appropriate language depending on the \texttt{lang\_pair} attribute, which \textsc{fairseq} dynamically sets, and is guaranteed to match that of the input.

\begin{listing*}
\begin{minted}[mathescape,
               linenos,
               %numbersep=5pt,
               gobble=0,
               fontsize=\scriptsize,
               frame=lines,
               framesep=2mm,
               breaklines, 
               breakafter=d,
               numbersep=-5pt,
               ]{python}
    class LanguageSpecificEncoderLayer(nn.Module):
      def __init__(self, args, layer=0):
          super().__init__()
          self.index_language = args.language_specific_layers[layer]
          all_languages = sorted(set(self.get_lang(lp) for lp in args.lang_pairs))  
          self.models = nn.ModuleDict({lang: TransformerEncoderLayer(args, layer) for lang in all_languages})

      def get_lang(self, lang_pair):
          # lang_pair is, for example: "en_US-de_DE"  
          if self.index_language == "src":
              return lang_pair.split("-")[0]
          elif self.index_language == "tgt":
              return lang_pair.split("-")[1]
          else:
              raise ValueError(f"Invalid language `{self.index_language}`.")

      def forward(self, x, encoder_padding_mask, attn_mask: Optional[Tensor] = None):
          # self.lang_pair is set dynamically from outside the module.
          return self.models[self.get_lang(self.lang_pair)].forward(x, encoder_padding_mask, attn_mask)
\end{minted}
\caption{Sample implementation of a \ourslong in \textsc{fairseq}.}
\label{code:mtl}
\end{listing*}

\section{Dataset sizes}
\label{app:dataset_sizes}
For most language pairs, CCMatrix is the largest data source, and it is also the lowest quality one. To compensate for this quality imbalance, we apply temperature sampling \citep{DBLP:journals/corr/abs-1907-05019} to balance the different sources, using a temperature of $5$, which worked well in our experiments. In our initial experiments, we considered two approaches to apply this temperature re-sampling: either upsampling the higher quality sources (WMT21, Opus-100, and Tatoeba), or downsampling CCMatrix. The results between these two approaches were similar, and since the downsampling runs were faster and more stable, we used the downsampling for all our experiments. To avoid discarding too much data, we capped the maximum downsampling to a factor of $10$.

\Cref{tab:dataset_sizes} shows the number of sentence pairs for each language direction, after de-duplication, cleaning, and downsampling CCMatrix.

\begin{table*}
    \centering
    \small
    \begin{tabular}{cccccccccc}
    \toprule 
    & \len & \les & \lfr & \lit & \lja & \lko & \lpt & \lsw & \lzh \\
    \midrule
    \lde & $213$M & $11.6$M & $36.6$M & $7.2$M & $1.2$M & $708$K & $5.4$M & $2.4$M & $1.8$M \\
    \len & $-$ & $230$M & $286$M & $96.3$M & $36.5$M & $2.3$M & $78.6$M & $708$K & $88.9$M \\
    \les & $-$ & $-$ & $49.4$M & $14.9$M & $1.3$M & $772$K & $22.3$M &  $6.9$M &  $6.9$M \\
    \lfr & $-$ & $-$ & $-$ & $14.9$M & $1.2$M & $752$K & $12.9$M & $8$M & $25.3$M \\
    \lit & $-$ & $-$ & $-$ & $-$ & $736$K & $382$K & $7$M &  $1.1$M & $964$K \\
    \lja & $-$ & $-$ & $-$ & $-$ & $-$ & $511$K & $764$K &  $820$K & $897$K \\
    \lko & $-$ & $-$ & $-$ & $-$ & $-$ & $-$ & $756$K & $536$K & $3$M \\
    \lpt & $-$ & $-$ & $-$ & $-$ & $-$ & $-$ & $-$ & $3.6$M & $1.1$M \\
    \lsw & $-$ & $-$ & $-$ & $-$ & $-$ & $-$ & $-$ & $-$ & $962$K \\
    \bottomrule 
    \end{tabular}
    \caption{Number of training sentence pairs for each language pair, after data de-duplication, cleaning, and downsampling CCMatrix. We report only one language direction, as the data is the same for both directions.}
    \label{tab:dataset_sizes}
\end{table*}

\section{Full results}
\label{app:full_results}
\Cref{tab:full_results} shows the \chrf scores on the Flores-101 test set for all language directions, of both our shared-encoder, separate-decoder baseline model and our proposed \oursnas architecture with pre-training. Statistically non-significant results ($p \geq 0.05$) are marked with $\nosig$ (in total $6$ of the $90$ language pairs). The results on the validation set can be found in \Cref{tab:full_results_valid}.

\begin{table*}
\centering
{\setlength{\tabcolsep}{2.4pt}
    \begin{minipage}{.47\textwidth}
    \scriptsize
    \centering
    \begin{tabular}{p{6pt}lLLLLLLLLLL}
    \toprule 
    & & \multicolumn{1}{c}{\lde} & \multicolumn{1}{c}{\len} & \multicolumn{1}{c}{\les} & \multicolumn{1}{c}{\lfr} & \multicolumn{1}{c}{\lit} & \multicolumn{1}{c}{\lja} & \multicolumn{1}{c}{\lko} & \multicolumn{1}{c}{\lpt} & \multicolumn{1}{c}{\lsw} & \multicolumn{1}{c}{\lzh} \\
    \midrule
    \lde & $\rightarrow$ & - & 67.4 & 52.1 & 61.0 & 54.2 & 30.1 & 26.1 & 59.0\nosig & 48.1 & 31.1 \\
    \len & $\rightarrow$ & 64.6 & - & 55.6 & 70.7 & 58.5\nosig & 35.0 & 28.8 & 70.6 & 55.4 & 34.9 \\
    \les & $\rightarrow$ & 53.1 & 59.2 & - & 57.7 & 52.6 & 28.0 & 23.4 & 54.7\nosig & 47.7 & 27.7 \\
    \lfr & $\rightarrow$ & 57.8 & 68.3 & 52.7 & - & 55.6 & 31.1 & 25.9 & 60.5\nosig & 50.1\nosig & 31.1 \\
    \lit & $\rightarrow$ & 54.9 & 61.4 & 52.3 & 60.3 & - & 28.8 & 24.7 & 56.6 & 49.2 & 29.7 \\
    \lja & $\rightarrow$ & 46.1 & 53.3 & 44.2 & 50.0 & 45.1 & - & 26.1 & 47.5 & 42.3 & 25.2 \\
    \lko & $\rightarrow$ & 43.5 & 49.6 & 41.2 & 46.6 & 42.0 & 28.0 & - & 45.3 & 40.5 & 23.4 \\
    \lpt & $\rightarrow$ & 58.6 & 71.1 & 53.8\nosig & 64.0 & 55.5 & 30.5 & 26.4 & - & 53.1 & 31.8 \\
    \lsw & $\rightarrow$ & 47.0 & 57.3 & 43.6 & 51.0 & 44.5 & 23.0 & 20.5 & 50.0 & - & 23.9 \\
    \lzh & $\rightarrow$ & 48.2 & 56.3 & 46.3 & 52.7 & 47.0 & 26.8 & 24.5 & 49.6 & 44.3 & - \\
    \bottomrule 
    \end{tabular}
    \end{minipage}
    \quad
    \begin{minipage}{.47\textwidth}
    \scriptsize
    \centering
    \begin{tabular}{p{6pt}lLLLLLLLLLL}
    \toprule 
    & & \multicolumn{1}{c}{\lde} & \multicolumn{1}{c}{\len} & \multicolumn{1}{c}{\les} & \multicolumn{1}{c}{\lfr} & \multicolumn{1}{c}{\lit} & \multicolumn{1}{c}{\lja} & \multicolumn{1}{c}{\lko} & \multicolumn{1}{c}{\lpt} & \multicolumn{1}{c}{\lsw} & \multicolumn{1}{c}{\lzh} \\
    \midrule
    \lde & $\rightarrow$ & - & 67.9 & 52.8 & 62.2 & 55.0 & 32.2 & 28.0 & 59.1\nosig & 49.2 & 32.0 \\
    \len & $\rightarrow$ & 65.1 & - & 56.2 & 71.2 & 58.6\nosig & 37.9 & 30.0 & 71.1 & 56.6 & 35.9 \\
    \les & $\rightarrow$ & 54.0 & 59.7 & - & 58.3 & 52.9 & 29.5 & 25.2 & 54.9\nosig & 49.1 & 29.1 \\
    \lfr & $\rightarrow$ & 58.6 & 68.9 & 53.0 & - & 56.2 & 32.0 & 28.0 & 60.8\nosig & 50.4\nosig & 32.1 \\
    \lit & $\rightarrow$ & 55.7 & 61.7 & 53.0 & 60.7 & - & 29.9 & 26.4 & 57.1 & 50.6 & 30.2 \\
    \lja & $\rightarrow$ & 47.9 & 55.0 & 45.1 & 50.6 & 45.9 & - & 27.3 & 48.6 & 44.4 & 26.7 \\
    \lko & $\rightarrow$ & 45.4 & 52.1 & 42.9 & 48.2 & 43.9 & 30.7 & - & 47.2 & 43.2 & 25.6 \\
    \lpt & $\rightarrow$ & 59.5 & 71.5 & 54.0\nosig & 64.7 & 55.9 & 31.6 & 28.9 & - & 54.6 & 32.7 \\
    \lsw & $\rightarrow$ & 49.2 & 59.7 & 45.2 & 52.9 & 46.1 & 25.8 & 22.7 & 52.0 & - & 24.9 \\
    \lzh & $\rightarrow$ & 49.7 & 57.1 & 47.3 & 53.9 & 47.9 & 29.1 & 26.5 & 50.0 & 46.5 & - \\
    \bottomrule 
    \end{tabular}
    \end{minipage}
    \caption{Comparison of the baseline model (\emph{left}) and our learned architecture \oursnas with dense pre-training (\emph{right}) for each language pair, on the Flores-101 test set. Our approach gives \emph{significant} \chrf gains for most language pairs. Statistically \insig improvements using paired bootstrap resampling are marked with $\nosig$ for $p \geq 0.05$ (in total $6$ of the $90$ language pairs).}
    \label{tab:full_results}
    }
\end{table*}

\begin{table*}
\centering
{\setlength{\tabcolsep}{2.4pt}
    \begin{minipage}{.47\textwidth}
    \scriptsize
    \centering
    \begin{tabular}{p{6pt}lLLLLLLLLLL}
    \toprule 
    & & \multicolumn{1}{c}{\lde} & \multicolumn{1}{c}{\len} & \multicolumn{1}{c}{\les} & \multicolumn{1}{c}{\lfr} & \multicolumn{1}{c}{\lit} & \multicolumn{1}{c}{\lja} & \multicolumn{1}{c}{\lko} & \multicolumn{1}{c}{\lpt} & \multicolumn{1}{c}{\lsw} & \multicolumn{1}{c}{\lzh} \\
    \midrule
    \lde & $\rightarrow$ & - & 67.4 & 50.9 & 60.6 & 53.6 & 31.6 & 26.2 & 58.5 & 48.3 & 30.3 \\
    \len & $\rightarrow$ & 64.0 & - & 55.4 & 70.8 & 58.4 & 35.7 & 29.0 & 70.0 & 55.5 & 33.3 \\
    \les & $\rightarrow$ & 52.4 & 59.7 & - & 57.6 & 52.6 & 28.5 & 24.0 & 54.2 & 48.4 & 27.8 \\
    \lfr & $\rightarrow$ & 57.6 & 68.8 & 52.3 & - & 55.5 & 30.6 & 25.5 & 60.0 & 50.1 & 29.7 \\
    \lit & $\rightarrow$ & 54.3 & 61.8 & 51.4 & 59.7 & - & 29.6 & 24.0 & 55.8 & 49.1 & 29.1 \\
    \lja & $\rightarrow$ & 46.0 & 54.0 & 43.5 & 49.3 & 45.4 & - & 26.4 & 47.2 & 42.4 & 25.2 \\
    \lko & $\rightarrow$ & 43.1 & 50.1 & 40.6 & 45.9 & 42.2 & 27.8 & - & 44.7 & 40.3 & 22.7 \\
    \lpt & $\rightarrow$ & 57.9 & 71.2 & 53.1 & 63.8 & 54.9 & 31.0 & 26.9 & - & 52.5 & 30.9 \\
    \lsw & $\rightarrow$ & 47.2 & 58.4 & 43.7 & 50.9 & 44.8 & 23.9 & 20.7 & 50.3 & - & 23.9 \\
    \lzh & $\rightarrow$ & 48.0 & 56.6 & 45.9 & 52.4 & 47.0 & 27.6 & 24.5 & 49.0 & 44.2 & - \\
    \bottomrule 
    \end{tabular}
    \end{minipage}
    \quad
    \begin{minipage}{.47\textwidth}
    \scriptsize
    \centering
    \begin{tabular}{p{6pt}lLLLLLLLLLL}
    \toprule 
    & & \multicolumn{1}{c}{\lde} & \multicolumn{1}{c}{\len} & \multicolumn{1}{c}{\les} & \multicolumn{1}{c}{\lfr} & \multicolumn{1}{c}{\lit} & \multicolumn{1}{c}{\lja} & \multicolumn{1}{c}{\lko} & \multicolumn{1}{c}{\lpt} & \multicolumn{1}{c}{\lsw} & \multicolumn{1}{c}{\lzh} \\
    \midrule
    \lde & $\rightarrow$ & - & 67.9 & 51.1 & 61.2 & 54.3 & 32.9 & 27.8 & 58.3 & 48.4 & 31.5 \\
    \len & $\rightarrow$ & 65.1 & - & 55.5 & 71.1 & 58.9 & 37.8 & 31.0 & 70.4 & 57.2 & 34.5 \\
    \les & $\rightarrow$ & 53.5 & 59.9 & - & 58.2 & 52.8 & 29.4 & 25.7 & 54.4 & 49.3 & 28.8 \\
    \lfr & $\rightarrow$ & 58.4 & 69.5 & 52.4 & - & 56.1 & 32.5 & 28.0 & 59.9 & 50.1 & 30.9 \\
    \lit & $\rightarrow$ & 55.1 & 62.0 & 51.7 & 60.0 & - & 30.6 & 26.8 & 56.1 & 50.5 & 30.1 \\
    \lja & $\rightarrow$ & 47.1 & 55.2 & 44.5 & 51.1 & 46.3 & - & 27.5 & 48.5 & 44.3 & 26.2 \\
    \lko & $\rightarrow$ & 45.1 & 52.2 & 42.0 & 47.6 & 43.6 & 30.7 & - & 45.8 & 43.0 & 24.6 \\
    \lpt & $\rightarrow$ & 58.8 & 71.7 & 53.4 & 64.6 & 55.3 & 32.5 & 29.3 & - & 53.9 & 31.8 \\
    \lsw & $\rightarrow$ & 49.0 & 61.0 & 45.0 & 52.9 & 46.5 & 26.4 & 23.7 & 52.6 & - & 24.2 \\
    \lzh & $\rightarrow$ & 49.2 & 57.7 & 46.4 & 53.3 & 47.7 & 29.8 & 26.3 & 50.1 & 46.4 & - \\
    \bottomrule 
    \end{tabular}
    \end{minipage}
    \caption{Comparison of the baseline model (\emph{left}) and our learned architecture \oursnas with dense pre-training (\emph{right}) for each language pair, on the Flores-101 validation set.}
    \label{tab:full_results_valid}
    }
\end{table*}

\section{Results on different configurations}
\label{app:different_configs}
In \Cref{tab:all_adapter_results} we show the results of further experiments with Adapter Blocks. Besides encoder source language adapters (Language Adapters \textsc{Enc}) and source adapters in the bottom half of the encoder and target adapters in the top half (Language Adapters \textsc{Enc (Src+Tgt)}, we include source adapters on the encoder and target adapters on the decoder (Language Adapters \textsc{Enc+Dec}, like \citet{philip-etal-2020-monolingual}, and language-pair adapters on the encoder \citep{bapna-firat-2019-simple} (Language-Pair Adapters \textsc{Enc}). Our proposed architecture, \oursnas, outperforms all other techniques while introducing no extra computation at inference time (\ie, it keeps $|\boldsymbol{\theta}_{\text{eff}}|$ constant). 

\begin{table}[t]
    \centering
    \scriptsize
    {\setlength{\tabcolsep}{1.2pt}
    \begin{tabular}{lcccc}
    \toprule 
    \multicolumn{1}{l}{Model} & \chrf & \spBLEU & $|\boldsymbol{\theta}|$ & $|\boldsymbol{\theta}_{\text{eff}}|$ \\
    \midrule
    \multicolumn{1}{l}{Separate Decoder Baseline} & $45.5$ & $26.0$ & $299$M & $186$M \\
    \multicolumn{1}{l}{\oursnas} & $46.4$ & $27.2$ & $441$M & $186$M \\
    \multicolumn{1}{l}{\quad \emph{+ Dense Pre-training}} & $\mathbf{46.8}$ & $\mathbf{27.5}$ & $441$M & $186$M \\
    \midrule
    Language Adapters \textsc{Enc} $128$ & $45.8$ & $26.4$ & $321$M & $207$M \\
    \multicolumn{1}{l}{\quad + hidden dim $640$} & $46.2$ & $26.9$ & $426$M & $261$M \\
    Language Adapters \textsc{Enc} $256$ & $45.7$ & $26.3$ & $342$M & $228$M \\
    \multicolumn{1}{l}{\quad + hidden dim $640$} & $46.3$ & $27.1$ & $452$M & $282$M \\
    Language Adapters \textsc{Enc} $512$ & $45.6$ & $26.3$ & $384$M & $270$M \\
    \cdashlinelr{1-5}
    Language Adapters \textsc{Enc (Src+Tgt)} $128$ & $45.7$ & $26.3$ & $321$M & $207$M \\
    Language Adapters \textsc{Enc (Src+Tgt)} $256$ & $46.1$ & $26.7$ & $342$M & $228$M \\
    Language Adapters \textsc{Enc (Src+Tgt)} $512$ & $46.0$ & $26.7$ & $384$M & $270$M \\
    \cdashlinelr{1-5}
    Language-Pair Adapters \textsc{Enc} $128$ & $45.2$ & $25.7$ & $491$M & $207$M \\
    Language-Pair Adapters \textsc{Enc} $256$ & $45.3$ & $25.8$ & $680$M & $228$M \\
    Language-Pair Adapters \textsc{Enc} $512$ & $45.3$ & $25.9$ & $1057$M & $270$M \\
    \cdashlinelr{1-5}
    Language Adapters \textsc{Enc+Dec} $256$ & $45.5$ & $26.0$ & $350$M & $236$M \\
    Language Adapters \textsc{Enc+Dec} $512$ & $46.0$ & $26.4$ & $400$M & $286$M \\
    Language Adapters \textsc{Enc+Dec} $768$ & $46.1$ & $26.6$ & $449$M & $336$M \\
    Language Adapters \textsc{Enc+Dec} $1024$ & $46.2$ & $26.7$ & $499$M & $385$M \\
    \bottomrule 
    \end{tabular}}
    \caption{Comparison of different Adapter Blocks configurations on the separate decoder architecture.}
    \label{tab:all_adapter_results}
\end{table}

\section{Number of \oursm}
\label{app:number_of_lsls}
\begin{figure}[t]
    \centering
    \includegraphics[width=\columnwidth]{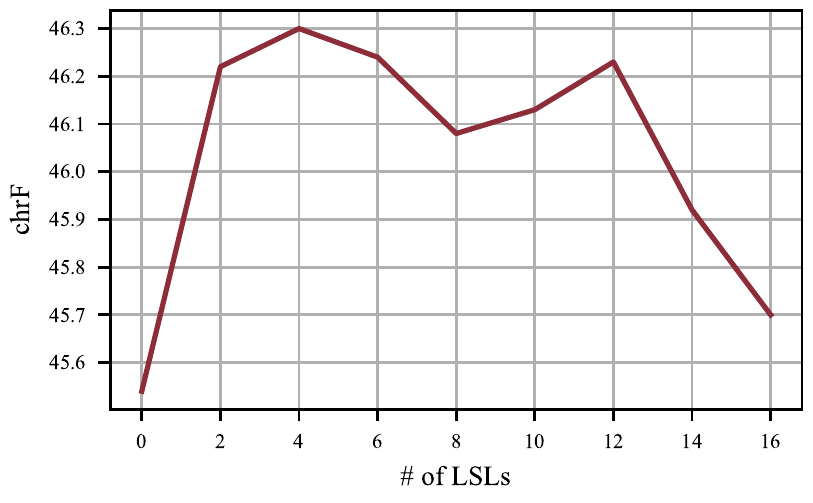}
    \caption{Average \chrf scores over all languages versus the number of \oursm. For each data point, half of the \oursm are on the bottom of the encoder, and the other half are on the top, \eg for $4$ \oursm, the bottom $2$ layers are source-indexed, the top $2$ are target-indexed, and the remaining layers are shared.}
    \label{fig:bleu-vs-layers}
\end{figure}

We look at the effect of changing the number of \oursm, illustrated in \Cref{fig:bleu-vs-layers}. To this end, we change the number of \oursm from $0$ to $16$, in increments of $2$, and, for each point, we place an additional \ours on the bottom and on the top of the encoder, using the source and target languages to index them, respectively. For example, $4$ \oursm corresponds to \arch{1,2}{15,16}. We see that adding more \oursm helps performance, but only up to a point (in this case, $4$ layers), and that afterwards, performance degrades, except for an outlier at $12$ \oursm. This implies that while the language-specific layers boost performance, having shared layers is crucial for knowledge transfer.

\section{Per Language results}
\label{app:per_lang_results}
\begin{table}[t]
    \centering
		\small
		\begin{tabularx}{0.78\columnwidth}{p{6pt}p{2pt}p{10pt}ccc}
    		\toprule
    		\multicolumn{3}{c}{Direction} & Sep. Decoder & Ours & $\Delta$ \\
    		\midrule
            \textsc{Eur} & $\rightarrow$ & \textsc{Eur} & $59.1$ & $59.7$ & \chunb{0.6} \\
            \textsc{Eur} & $\rightarrow$ & \textsc{CJK} & $29.2$ & $30.6$ & \chunb{1.4} \\
            \textsc{Eur} & $\rightarrow$ & \lsw & $50.6$ & $51.7$ & \chunb{1.1} \\
            \midrule
            \textsc{CJK} & $\rightarrow$ & \textsc{Eur} & $47.4$ & $48.8$ & \chunb{1.4} \\
            \textsc{CJK} & $\rightarrow$ & \textsc{CJK} & $25.7$ & $27.7$ & \chunb{2.0} \\
            \textsc{CJK} & $\rightarrow$ & \lsw & $42.4$ & $44.7$ & \chunb{2.3} \\
            \midrule
            \lsw & $\rightarrow$ & \textsc{Eur} & $48.9$ & $50.9$ & \chunb{2.0} \\
            \lsw & $\rightarrow$ & \textsc{CJK} & $22.4$ & $24.4$ & \chunb{2.0} \\
    		\bottomrule
		\end{tabularx}
		\caption{Comparison of \oursnas with pre-training compared to the separate baseline model per language family. Our approach gives substantial average \chrf gains for all, which are statistically significant for all but $6$ of the $90$ translation directions.}
		\label{tab:lang-group-results-mtl}
\end{table}
In \Cref{tab:lang-group-results-mtl}, we show aggregate scores for each language group: European (\lde, \len, \les, \lfr, \lit, \lpt), CJK (\lzh, \lja, \lko), and \lsw (isolated, since it is the only language in its family). Here, we see a similar trend, with our approach showing clear improvements both within groups, and between different groups.

\section{\comet results}
\label{app:comet_results}
We show \comet, \chrf, and \spBLEU{} scores, averaged over all language pairs in \Cref{tab:comet_results}. We show the scores for the baseline (\ie, non-\ours), our \ours model, and the best Adapter model for both the separate decoder and the shared decoder architectures. In all metrics, our proposed architectures outperform the remaining models.

\begin{table}[t]
    \centering
    \scriptsize
    {\setlength{\tabcolsep}{3pt}
    \begin{tabular}{lccc}
    \toprule 
    Model & \comet & \chrf & \spBLEU \\
    \midrule
    Separate Decoder Baseline & $0.45285$ & $45.5$ & $26.0$ \\
    \oursnas & $0.49577$ & $46.4$ & $27.2$ \\
    \multicolumn{1}{l}{\quad \emph{+ Dense Pre-training}} & $0.50759$ & $46.8$ & $27.5$ \\
    Language Adapters \textsc{Enc (Src+Tgt)} $256$ & $0.48265$ & $46.1$ & $26.7$ \\
    \cdashlinelr{1-4}
    Shared Decoder Baseline & $0.36975$ & $44.7$ & $24.9$ \\
    \oursnassd & $0.46542$ & $46.3$ & $26.7$ \\
    \multicolumn{1}{l}{\quad \emph{+ Dense Pre-training}} & $0.48357$ & $46.6$ & $27.1$ \\
    Shared Decoder Adapters $512$ & $0.41849$ & $45.3$ & $25.6$ \\
    \bottomrule 
    \end{tabular}}
    \caption{\comet, \chrf, and \spBLEU scores for the (non-\ours) baseline, our \ours models, and the best adapter model for the separate decoder and shared decoder architectures. These scores are averaged over all language pairs.}
    \label{tab:comet_results}
\end{table}

\section{Inference Speed}
\label{app:inference_speed}
We report the inference times for the various architectures we considered in \Cref{tab:inference_speed}. We report tokens/second on the \lde-\len test set\footnote{We repeated these measurements for language pairs, such as \len-\lzh, with similar results.}, averaged over $5$ runs. Our latency measurements were collected using a single NVIDIA V100 GPU (Speed GPU) or a single-threaded Intel Xeon Platinum 8275CL CPU @ 3.00GHz (Speed CPU), both with batch size of $1$, which faithfully captures the inference on a deployed neural machine translation model. As expected, the latency of shared decoder models is the same as that of similar separate decoder models (since only one of the decoders is used at inference time) so, for succinctness, we only report the separate decoder numbers.

A couple of comments regarding the Adapter models: 1) we do not report speed numbers for the ``Language Adapters \textsc{Enc (Src+Tgt)}'' as the architecture is the same as ``Language Adapters \textsc{Enc}''; 2) inference speed does not change significantly when adding encoder adapters, but only when adding adapters to the decoder.

\begin{table}[t]
    \centering
		\scriptsize
		\begin{tabularx}{\columnwidth}{lcc}
    		\toprule
    		Architecture & Speed GPU & Speed CPU \\
    		\midrule
            Shared Decoder Baseline & $195.2\pm2.6$ & $61.4\pm0.3$ \\
            Separate Decoder Baseline & $194.3\pm1.4$ & $61.7\pm0.2$ \\
            \multicolumn{1}{l}{\quad + hidden dim $640$} & $191.9\pm1.6$ & $54.0\pm0.2$ \\
            \multicolumn{1}{l}{\quad + hidden dim $704$} & $189.8\pm1.7$ & $51.6\pm0.3$ \\
            \multicolumn{1}{l}{\quad + hidden dim $768$} & $187.7\pm2.1$ & $48.4\pm0.2$ \\
            \midrule
            Language Adapters \textsc{Enc} $128$ & $188.1\pm1.8$ & $61.2\pm0.3$ \\
            Language Adapters \textsc{Enc} $256$ & $186.0\pm1.6$ & $61.1\pm0.3$ \\
            Language Adapters \textsc{Enc} $512$ & $187.6\pm1.1$ & $61.0\pm0.2$ \\
            \cdashlinelr{1-3}
            Language Adapters \textsc{Enc+Dec} $256$ & $165.2\pm2.4$ & $57.6\pm0.3$ \\
            Language Adapters \textsc{Enc+Dec} $512$ & $165.1\pm4.5$ & $57.2\pm0.2$ \\
            Language Adapters \textsc{Enc+Dec} $768$ & $164.4\pm2.1$ & $56.9\pm0.3$ \\
            \midrule
            \oursnas & $195.0\pm1.1$ & $61.3\pm0.2$ \\
            \oursnassd & $195.5\pm4.7$ & $61.4\pm0.3$ \\
    		\bottomrule
		\end{tabularx}
		\caption{Tokens/second comparison of different models on the Flores-101 \lde-\len test set. We show the average over $5$ runs, and the associated standard deviation. The latency of shared decoder models is the same as that of similar separate decoder models so, for succinctness, we only report the separate decoder numbers.}
		\label{tab:inference_speed}
\end{table}

\end{document}